\documentclass[letterpaper, 10 pt, conference]{ieeeconf}  

\IEEEoverridecommandlockouts                              

\usepackage{cite}
\usepackage{amsmath,amssymb,amsfonts}
\usepackage{graphicx}
\graphicspath{ {./images/} }
\usepackage{textcomp}
\usepackage{xcolor}
\usepackage{subfig}
\usepackage{url}
\usepackage{algpseudocode}
\usepackage{algorithm}
\usepackage{caption}
\usepackage{siunitx}
\DeclareCaptionLabelFormat{bold}{\textbf{Fig. #2. }}
\def\BibTeX{{\rm B\kern-.05em{\sc i\kern-.025em b}\kern-.08em
    T\kern-.1667em\lower.7ex\hbox{E}\kern-.125emX}}

\usepackage{float}

\title{\LARGE 
Newton-Raphson Flow for Aggressive Quadrotor Tracking Control
}

\author{Evanns Morales-Cuadrado, Christian Llanes, Yorai Wardi, and Samuel Coogan
\thanks{This work was supported in part by the NASA University Leadership Initiative (ULI) under grant 80NSSC20M0161 and by the National Science Foundation under grants \#1749357 and \#1924978.}
\thanks{The authors are with the School of Electrical and Computer Engineering at the Georgia Institute of Technology, \texttt{\{egm,christian.llanes,ywardi,sam.coogan\}@gatech.edu}. S. Coogan is also with the School of Civil and Environmental Engineering.}%
}

\begin{document}

\maketitle
\thispagestyle{empty}
\pagestyle{empty}

\newcommand{\desiredwidth}{0.6\textwidth} 

\begin{abstract}
We apply the Newton-Raphson flow tracking controller to aggressive quadrotor flight and demonstrate that it achieves good tracking performance over a suite of benchmark trajectories, beating the native trajectory tracking controller in the popular PX4 Autopilot. The Newton-Raphson flow tracking controller is a recently proposed integrator-type controller that aims to drive to zero the error between a future predicted system output and the reference trajectory. This controller is computationally lightweight, requiring only an imprecise predictor, and achieves guaranteed asymptotic error bounds under certain conditions. We show that these theoretical advantages are realizable on a quadrotor hardware platform. Our experiments are conducted on a Holybrox x500v2 quadrotor using a Pixhawk 6x flight controller and a Rasbperry Pi 4 companion computer which receives location information from an OptiTrack motion capture system and sends input commands through the ROS2 API for the PX4 software stack.


\end{abstract}

\section{Introduction}\label{sec:intro}
Recent years have seen powerful new quadrotor flight controllers like the Pixhawk Series \cite{PixhawkSeries_standards} reach the market, enabling fast and off-the-shelf development for researchers and hobbyists. 
Substantial research considers LQR control \cite{PID_LQR, LQR_quadrotor}, model predictive control \cite{MPC_Realtime_Quad}, and differential flatness approaches \cite{mellinger_controller1, agressive_nonlininv_diffflatness}. Robust control for quadrotors has also been explored using $H_{\infty}$ approaches in \cite{H_infty1, H_infinity_PID, backstepping_hinfinity}. Moreover, approaches which do not require linearization of the system dynamics such as backstepping have been explored in \cite{backstepping_hinfinity}, which also utilizes $H_{\infty}$ for disturbance rejection. However, while powerful, many of the controllers mentioned previously are computationally complex; require advanced knowledge of the reference input and its derivatives; or require detailed knowledge of the nonlinear dynamic model and its parameters. The result is that there emerges a practical state-of-the-art that is largely tethered to advanced PID control methodologies such as \cite{backstepping_Yusuf}, and is generally divorced from the true state-of-the-art in the research literature \cite{PID_Quad_Survey}.


\begin{figure}[t!]
    \centering

    \includegraphics[width=0.50\textwidth, height=0.28\textwidth]{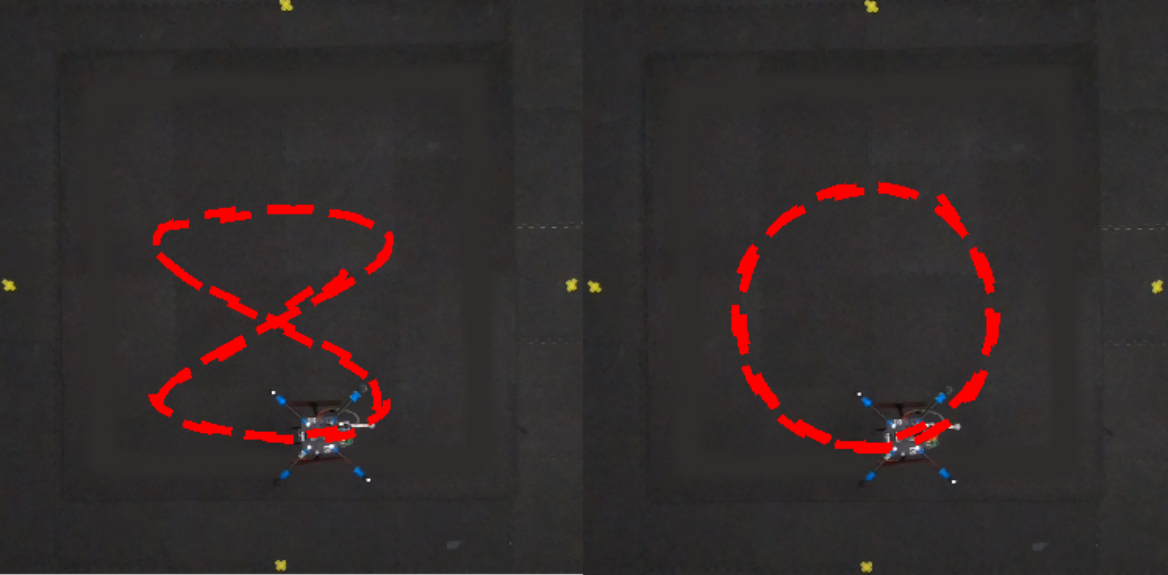}

    \caption{Frames from overhead videos of a quadrotor performing lemniscates and circles using the proposed Newton-Raphson flow controller with the trajectories overlaid.
}
    \label{fig:superimposed_frames}

    \vspace{-0.2in}

\end{figure}

In this paper, we present an integrator-type controller for aggressive quadrotor trajectory tracking. The controller has its roots in the Newton-Raphson method, an iterative numerical method for finding the zeros of a nonlinear function. Our control methodology requires a state predictor that approximates the state of the system at some future horizon $T$ given the system's current state and input. 
Then, a fluid-flow variant of the Newton-Raphson method updates the control input via a tunable integration update speed with the goal of driving to zero the error between the predicted future state and the reference trajectory. The result is a computationally lightweight feedback control scheme that achieves remarkable asymptotic tracking performance. In examples, we use a predictor based on linearized quadrotor dynamics and zero-order-input-hold, which requires only knowledge of the vehicle mass and gravitational constant and is comparable to the model information typical in PID-based control. We show that this simple predictor is
 remarkably effective, which emphasizes that the power of the proposed controller is in the combination of the Newton-Raphson flow and a reasonable, but not necessarily perfect, predictor.

One challenge of integrator-type control strategies is avoiding the integrator wind-up phenomenon that can lead to large transient overshoots in tracking performance or commanded inputs that exceed real-world actuator limits. Simple thresholding schemes can mitigate these issues but often result in abrupt changes to the inputs that degrade tracking performance. 
To mitigate these challenges, we incorporate an integral control barrier function into our controller architecture to keep control signals within acceptable limits based on the theory developed in \cite{ICBF}. The nonlinear barrier formulation prescribes the minimal alteration to the nominal Newton-Raphson flow required to achieve an acceptable control signal and, computationally, is easily applied as a post-processing step in the Newton-Raphson control update loop. The result is a controller that retains the desirable asymptotic tracking behavior of the Newton-Raphson controller while smoothly attenuating undesirable transients.

The proposed method is computationally comparable to PID control, requires limited knowledge of the true system dynamics, and includes few controller parameters, making it easy to tune. Indeed, the same parameter values have been shown to work well over a range of particular control applications. Lastly, we show that the proposed control strategy outperforms, sometimes dramatically, the standard PX4 trajectory tracking algorithm, which itself is a state-of-the-art cascaded PID feedback control architecture that uses a nonlinear mapping from desired accelerations to desired thrust and attitude based on geometric control techniques \cite{PX4_Controller_Diagrams}. 

There exists a body of work developing the Newton-Raphson tracking controller technique in various forms and guaranteeing asymptotic error bounds under certain conditions \cite{wardi_ijrnc_2019}.
This work is the first to successfully apply the Newton-Raphson flow controller to a real-world underactuated and open-loop unstable nonlinear plant, namely, the 6-DOF quadrotor, and we further demonstrate this novel application with a suite of hardware experiments\footnote{Our code and video demonstrations can be found at: \url{https://github.com/gtfactslab/MoralesCuadrado_ACC2024}}.



The rest of this paper is organized as follows. We introduce the continuous-time Newton-Raphson flow controller and its tracking error bounds in Section \ref{sec:newtonraphson derivation}. The quadrotor model and the predictor necessary for Newton-Raphson control are developed in Section \ref{sec:quadrotor model and predictor}. Section \ref{sec:CBFs} presents control barrier functions used to guarantee input safety limits. Hardware experiments are presented and analyzed in Section \ref{sec:experiments}. Finally, Section \ref{sec:conclusion} carries our closing remarks.

\section{Continuous-Time Newton-Raphson Tracking Controller}\label{sec:newtonraphson derivation}

The material presented in this section is contained in \cite{wardi_ijrnc_2019}.

\subsection{Newton-Raphson Flow Derivation}
The standard Newton-Raphson method for computing roots of algebraic functions $g:\mathbb{R}^m \to \mathbb{R}^m$ is derived from the first-order Taylor expansion as an iterative root-finding algorithm of the form 
\begin{equation}
    \label{eqn: NR}
\begin{aligned}
    u_n &= u_{n-1} - \left(\frac{dg}{du}(u_{n-1})\right)^{-1}g(u_{n-1}),
\end{aligned}
\end{equation}
$n=1,2,\ldots$, where the function $g(\cdot)$ is assumed to be continuously differentiable.
This simple iterative method can be reformulated as a Newton-Raphson output-tracking controller which aims at driving the error between a plant's output and a given reference target to zero. To do this, we restructure the algorithm with a memoryless plant relating the input $u$ to its corresponding output $y$ by the equation $y=g(u)$, to be run iteratively according to the formula
\begin{equation}
\label{eqn:iterative_NR_tracker}
\begin{aligned}
    u_{n} &= u_{n-1} + \left(\frac{\partial g}{\partial u}(u_{n-1})\right)^{-1}\big(r-g(u_{n-1})\big).  \\
\end{aligned}
\end{equation}
To place this method in a temporal context of continuous time, consider $u(t)$ and $y(t)$ as $t$-dependent quantities for all $t\geq 0$. Suppose that the control algorithm (2) executes a step every $\Delta t$ seconds for a given $\Delta t>0$. Define $u_{n}=u(n\Delta t)$, $n=0,1,\ldots$, subtract $u_{n-1}$ from both sides of (2), scale the resulting Right-Hand Side (RHS) of (2) by $\Delta t$, and take $\Delta t\to 0$. This gives the following differential equation for the control $u(t)$,
\begin{equation}
\label{eqn:continuous_time_memoryless_NR}
\begin{aligned}
\dot{u}(t) &= \left(\frac{\partial g}{\partial u}\bigl(u(t)\bigr)\right)^{-1} \bigl(r - g(u(t)\bigr).\\
\end{aligned}
\end{equation}
Eq. (3) defines the Newton-Raphson flow of Eq. (2). Under fairly general assumptions, it provably drives the error $r-y(t)$ to zero \cite{wardi_NR_first_prelim}.
In the subsequent discussion we expand this tracking controller in two ways: the target is no longer a constant but a time-dependent function $r(t)$, and the plant subsystem is not a memoryless nonlinearity but a dynamical system modeled by an ordinary differential equation. 

\subsection{Tracking Controller for Dynamical System with Time-dependent Reference Target}
Consider a dynamical system with the state equation 
\begin{equation}
\label{eqn:general nonlinear dynamical system}
\dot{x}(t) = f(x(t),u(t)), \hspace{1cm} t\geq 0
\end{equation}
and the output equation
\begin{equation}
y(t) = h(x(t)),
\end{equation}
with the state variable $x(t) \in \mathbb{R}^{n}$,
input $u(t) \in \mathbb{R}^{m}$ and output $y(t) \in \mathbb{R}^{m}$. Note that both the input $u$ and the output $y$ must be of the same dimension. The function $f(x,u)$ is assumed to be continuous in $(x,u)$, continuously differentiable in $x$ for a given $u$, and to satisfy standard sufficient conditions for a unique, continuous, piecewise-continuously differentiable solution $x(t)$, $t\in[0,\infty)$ 
for a given initial condition $x_{0}\in\mathbb{R}^n$ and every bounded, piecewise-continuously differentiable input trajectory $u(t)$ such as assumed in \cite{wardi_ijrnc_2019}. Also assume that the function $h(x)$ is continuously differentiable. The reference trajectory $r(t)\in\mathbb{R}^{m}$ is assumed to be bounded, continuous, and piecewise-continuously differentiable in $t\in[0,\infty)$.

Due to the dynamic nature of the system, it is no longer true that $x(t)$ hence $y(t)$ are functions only of $u(t)$ as in the aforementioned memoryless case. However, it is true that for a given $T>0$ and $t\in\mathbb{R}$, $x(t+T)$ hence $y(t+T)$ are functions of $x(t)$ and $\{u(\tau)$~:~ $\tau\in[t,t+T]\}$. But the future outputs $u(\tau)$, $\tau\in(t,t+T]$ cannot be assumed to be known at time $t$, therefore we predict the output $y(t+T)$ at time $t$ for the sake of extending the Newton-Raphson controller from Eq. (3) to the current dynamic setting. Such an output predictor is by no-means unique, and it is reasonable to require it to be functionally dependent on $(x(t),u(t))$. Formally, the predicted output, denoted by $\hat{y}(t+T)$, has the form
\begin{equation}
    \hat{y}(t+T)=\rho(x(t),u(t))
\end{equation} 
for a suitable function $\rho:\mathbb{R}^{n}\times\mathbb{R}^m \to \mathbb{R}^m$,
and we only require that the function $\rho$ be continuously differentiable in $(x(t),u(t))$. It need not have an analytic expression but has to be computable by simulation or numerical means. We label the function $\rho$ the {\it prediction function}. 

With this prediction function we extend the controller from (3) to the following equation,
\begin{align}
\label{eqn:continuous_time_dynamical_final_NR}
\dot{u}(t) &=\nonumber \\
&\alpha \left(  \frac{\partial \rho}{\partial u}\left(x(t),u(t)\right) \right)^{-1} \biggl(r(t+T) - \rho\bigl(x(t),u(t)\bigr)\biggr),
\end{align}
a word on the role of $\alpha$ will be said shortly. 
Equations (4) and (7) define the closed-loop system.

A practical choice of the predictor is to balance precision with computational efficiency, where the importance of the latter is in the fact that $\rho(x(t),u(t))$ has to be computed in real time as a part of the feedback law. In the literature of the Newton-Raphson flow method, both model-based \cite{wardi_ijrnc_2019} and neural-network based predictors \cite{NN_Pred_1} and \cite{NN_Pred_2} have been used.
In this paper we err on the side of computational efficiency by proposing a predictor that is computable by an explicit formula and has no need for on-line solution of a differential equation. The predictor function is based on the linearized dynamics of our system with a zero-order hold on the input for the duration of the lookahead period of $\tau \in [t, t+T]$. To the best of our knowledge, this is the first time the Newton-Raphson fluid flow tracking controller has been paired with a closed-form predictor for nonlinear systems.

The role of large $\alpha$ in Eq. (7), called the {\it controller's speedup}, is twofold: possibly stabilizing the closed-loop system if need be,
 and reducing the asymptotic tracking error. To explain the latter point, 
define the asymptotic prediction error by 
$\nu_{1}:=\lim\sup_{t\rightarrow\infty}||\hat{y}(t)-y(t)||$, define $\nu_{2}:=\lim\sup_{t\rightarrow\infty}||\dot{r}(t)||$, and denote the asymptotic tracking error by $\lim\sup_{t\rightarrow\infty}||r(t)-y(t)||$. According to \cite{wardi_ijrnc_2019}, under certain stability conditions, 
\begin{equation}
\label{eqn: output_convergence}
    \lim\sup_{t\rightarrow\infty} ||r(t) - y(t)|| \leq  \nu_1 +  \frac{\nu_2}{\alpha}.
\end{equation}
Moreover, \cite{wardi_ijrnc_2019} shows that other errors and system disturbances can be lumped with $\nu_{2}$ in (8) thereby reducing their corresponding upper bound on the asymptotic tracking error by a factor of $\alpha^{-1}$. Despite Eq. (8) demonstrating that asymptotic prediction error does not display such a behavior, we choose a predictor that emphasizes computational efficiency perhaps at the expense of prediction error and still obtain competitive results with commonly used, established predictors.

\section{Quadrotor Model and Predictor}\label{sec:quadrotor model and predictor}
The Newton-Raphson flow controller requires  a predictor for the system. In this section, we propose a simple predictor constructed from a small-angle approximation of the quadrotor dynamics. The motivation for choosing this predictor is that this simplified modeling assumption is common for tuning PID controllers and for developing controllers appropriate for nonaggressive tracking \cite{PID_Quad_Survey}. As we will see below, even with this imperfect predictor, we achieve notable tracking results. A full comparison of possible predictor methods, including those using high fidelity nonlinear or neural network models, is the subject of ongoing work. 

We take as the system state 
\begin{align}
{x} &= \begin{bmatrix}
        p_x & p_y & p_z & V_x & V_y & V_z & \phi & \theta & \psi
    \end{bmatrix}^T\in\mathbb{R}^9
\end{align}
where $(p_x,p_y,p_z)$ is the quadrotor's 3D position in a fixed world frame, $(V_x,V_y,V_z)$ is the quadrotor velocity in this frame, and ($\phi, \theta, \psi)$ are the quadrotor roll, pitch, and yaw angles. 

The system input is $u=\begin{bmatrix}u_\tau&u_p&u_q&u_r\end{bmatrix}^T\in\mathbb{R}^4$, consisting of three angular velocities $(u_p,u_q,u_r)$ and net thrust $u_\tau$ through the center of mass in the direction perpendicular to the plane of the rotors. The PX4 control architecture includes an off-board control mode that converts these inputs directly to commanded rotor velocities for the four rotor motors.

We take as the system output 
\begin{align}
y=Cx=\begin{bmatrix}p_x&p_y&p_z&\psi\end{bmatrix}^T\in\mathbb{R}^4
\end{align} for appropriately defined binary matrix $C\in\mathbb{R}^{4\times 9}$ (the detailed structure of $C$ is provided in the Appendix).

\subsection{Simplified Predictor Dynamics}\label{subsec:simplified predictor dynamics}
Even though we wish to control the quadrotor in aggressive regimes, to build a computationally simple predictor, we approximate the dynamics with a classical small-angle approximation and linearize at hover. The resulting dynamics used for the predictor are
\begin{align}
\label{simpleQuadDyn1}
(\dot{p}_x,\dot{p}_y,\dot{p}_z)&=(V_x,V_y,V_z)\\
\label{simpleQuadDyn2}
(\dot{\phi},\dot{\theta},\dot{\psi})&=(u_p,u_q,u_r)\\
\label{simpleQuadDyn3}
(\dot{V}_x,\dot{V}_y,\dot{V}_z)&= \left(-g\phi, g\theta, \frac{u_\tau}{m}-g\right)
\end{align}
where $m$ is the quadrotor mass and $g$ is the acceleration due to gravity. The simplified dynamics for prediction are linear and of the form $\dot{{x}}=A{x}+Bu$ for appropriate matrices $A$ and $B$, where the matrices $A$ and $B$ are defined in the Appendix.

Given state ${x}(t)$ and input $u(t)$ at time $t$, we take as the predicted state at time $t+T$ the solution to the simplified dynamics 
\begin{align}
    {x}(t+T)&=e^{AT}{x}(t) +\int _{0}^T \, e^{ A(T-\tau)}B u(t)d\tau  \\
\label{eqn:predictor2}
    &= \Tilde{A}{x}(t) + \Tilde{B} u(t)
\end{align}
for appropriate $\Tilde{A}$ and $\Tilde{B}$ that depend only on $T$ and are computable offline in closed-form. Note that, in constructing a predictor, we do not know the value of $u(\tau)$ during the duration of the lookahead period $\tau \in [t,t+T]$, and instead only know its value at the beginning of the period $\tau=t$. Therefore, we adopt a zero-order hold on the input, assuming it remains constant over the prediction horizon. We therefore obtain the final form for the predictor as
\begin{align}
\label{eq:rho}
    \hat{y}(t+T)=\rho(x(t),u(t))=C\tilde{A}x(t)+C\tilde{B}u(t)
\end{align}
where $\tilde{A}$, $\tilde{B}$, and $C$ are fixed throughout and computed as described above.

\subsection{Control Constraints via Integral Control Barrier Functions}\label{sec:CBFs}


In this section, we summarize and build on recent work proposing instead to use integral control barrier functions (I-CBFs), a variant of the popular control barrier function method enforcing forward invariance.
As originally developed in \cite{OG_CBFs}, control barrier functions provide a method to guide control inputs from leading to potentially unsafe states towards the most similar safe input. Integral CBFs, proposed in \cite{ICBF}, extend classical CBFs to the case of dynamically defined control laws, that is, control laws that update the rate of change of the input $\dot{u}$ rather than the input $u$ directly.

In our setting, we apply an I-CBF to mitigate integrator wind-up and transient overshoot as follows.

We first rewrite \eqref{eqn:continuous_time_dynamical_final_NR} as $\dot{u}(t) = \Psi(x(t),u(t),t)$ for appropriately defined $\Psi$. We then proceed by instead considering the modified 
dynamically defined control law
\begin{equation}
    \label{I-CBF general control law}
    \dot{u}(t) = \Psi(x(t),u(t),t) + \eta(t)
\end{equation}
where $\eta(t)$ is the minimal intervention term designed below to achieve desired performance. We consider a given scalar-valued barrier function $b(x,u)$ of state $x$ and input $u$ defining allowed state-input pairs: $b(x,u)\geq 0$ indicates that input $u$ is allowed at state $x$, while $b(x,u)<0$ implies input $u$ is not allowed at state $x$.

Then, following the methodology of \cite{ICBF}, we insist that $\eta(t)$ be chosen so that
\begin{equation}
\label{iCBF equation}
    \dot{b}(x(t),u(t),\eta(t)) + \gamma (b(x(t),u(t)) \geq 0
\end{equation}
for some fixed increasing and Lipschitz function $\gamma$ satisfying $\gamma(0)=0$, where 
\begin{align}
 \dot{b}(x,u,\eta)=\frac{\partial b}{\partial x}(x,u)^Tf(x,u)+\frac{\partial b}{\partial u}(x,u)^T(\Psi(x,u,t)+\eta).
\end{align}


As was developed in \cite{ICBF}, we can take in particular 
\begin{equation}
    \eta(t) =
    \begin{cases}
      \frac{\lambda(t)}{||\xi(t)||^2} \xi(t), & \text{if}\ \lambda(t) > 0 \\
      0, & \text{if}\ \lambda(t) \leq 0
    \end{cases}
\end{equation}
where
\begin{align}
\xi(t) &= \left( \frac{\partial b}{\partial u} (x(t),u(t)) \right)^{T } \\
\lambda(t) &= -\Bigg(\frac{\partial b}{\partial x} (x(t),u(t)) f(x(t),u(t)) \\
&+          \frac{\partial b}{\partial u} (x(t),u(t))  \Psi(x(t),u(t),t) + \gamma (b(x(t),u(t))) \Bigg).
\end{align}

\begin{algorithm}
\caption{Newton-Raphson Flow and I-CBFs for Rate Control}
\label{alg:NR_iCBF}

\textbf{Input:}
$x(t)=$ current system state, 
$u(t)=$current input, $r(t+T)=$ reference input $T$-units in the future, $\Delta t=$ controller update period.

\textbf{Initialize:} 

\texttt{RateMin}=$-0.8$, \texttt{RateMax}=$0.8$, \Comment{angular rate limits}

$\gamma=1.0$, \Comment{CBF parameter}

$\alpha=30, T=0.8$, \Comment{Newton-Raphson control parameters}


\textbf{Output:} $u(t+\Delta t)$, \Comment{input to be applied at next controller update step}

\begin{algorithmic}
\\
\State $(u_\tau,{u}_p,{u}_q,{u}_r)\gets u(t)$\Comment{Current input components}
$\begin{aligned}
(\tilde{\dot{u}}_{\tau},\tilde{\dot{u}}_p,&\tilde{\dot{u}}_q, \tilde{\dot{u}}_r)\gets \left(\frac{\partial \rho}{\partial u}\left(x(t),u(t)\right)\right)^{-1} \biggl(r(t+T) - \nonumber \\
    &\rho\bigl(x(t),u(t)\bigr)\biggr)
\end{aligned}$\\\Comment{$\rho$ from \eqref{eq:rho}}

\\
\For{$s\in\{p,q,r\}$}
        \If{${u}_s \geq 0$}
            \State $\eta\gets \min\{\gamma \cdot (\texttt{rateMax} - {u}_s) - \tilde{\dot{u}}_s,0\}$
            
        \Else{}
            \State $\eta \gets \max\{-\gamma \cdot (-\texttt{rateMin} - u_s) - \tilde{\dot{u}}_s,0\}$
        \EndIf
        \State $\dot{u}_s\gets \tilde{\dot{u}}_s+\eta$ \Comment{Apply integral control barrier to each commanded rate}
        \EndFor
        \State $\dot{u}\gets\begin{bmatrix}\dot{u}_\tau&\dot{u}_p&\dot{u}_q&\dot{u}_r\end{bmatrix}^T$
        \State $u(t+\Delta t)=u(t)+ \alpha \cdot \dot{u} \cdot \Delta t$  
\end{algorithmic}
\end{algorithm}

\section{Hardware Experiments}\label{sec:experiments}
\begin{figure*}[t]
    \centering

    \includegraphics[width=1.0\textwidth]{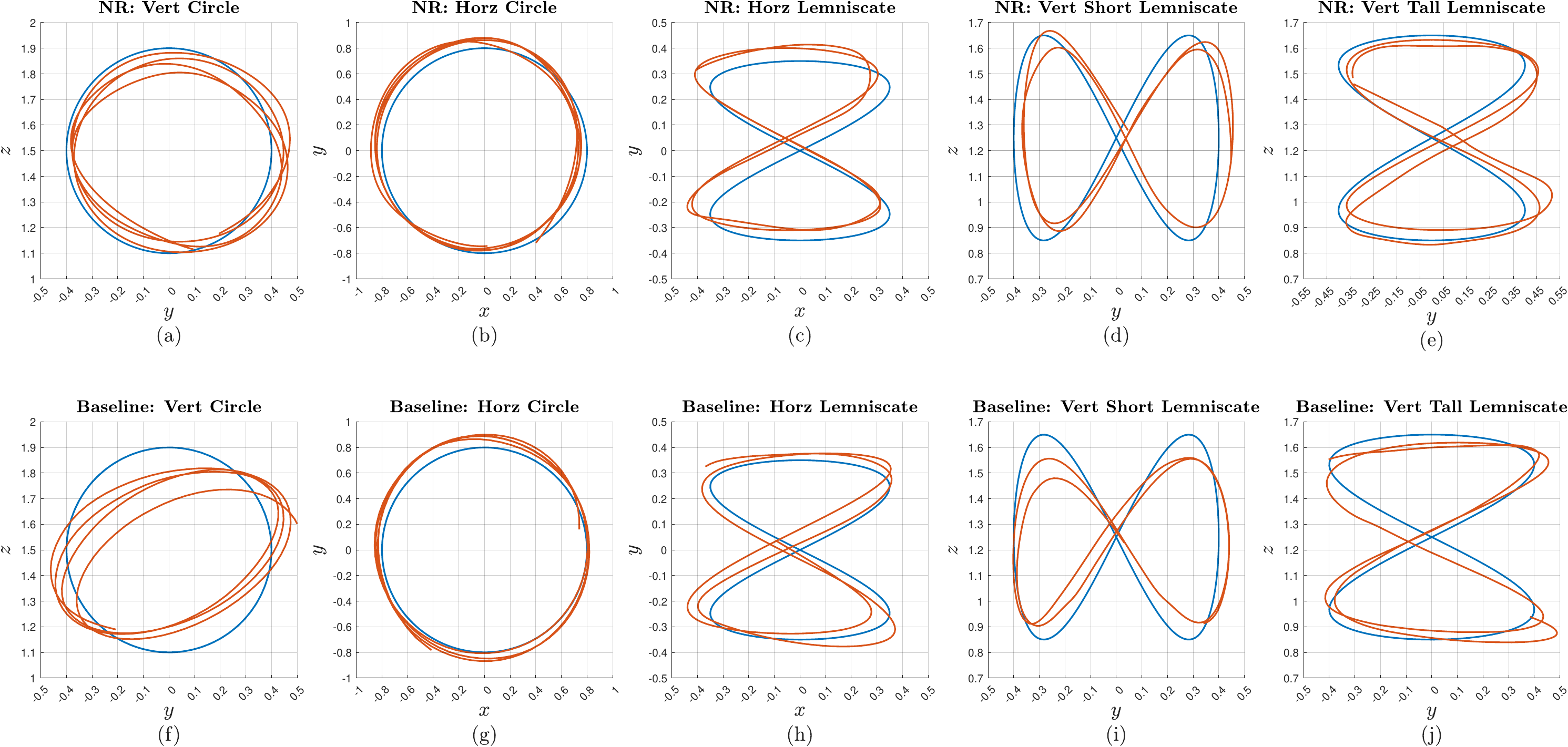}

    \caption{Comparison of five flight trajectories (red) with the reference trajectories (blue). Top row shows results using the Newton-Raphson flow control and the bottom row shows results using the baseline PX4 controller. Axes units are meters. The periods of circular and lemniscate trajectories are $3.14$ and $6.28$ seconds, respectively. In all cases, the Newton-Raphson flow controller outperforms the baseline controller,  evaluated with the root-mean-square error, as reported in Table \ref{tab:RMS_Comp}.}
    \label{fig:flight_trajecories_together}


\end{figure*}

We test the proposed Newton-Raphson flow tracking controller with a Holybro x500v2 quadrotor UAV with a Holybro Pixhawk 6X flight controller on a mini-base port. Onboard the quadrotor, we installed a Raspberry Pi 4 Model B which is responsible for running the algorithm and publishing commands to the flight controller through serial connection. SSH protocol is used to interact with this onboard computer. Note that we first test this controller in the Gazebo simulator with
advanced DART physics engine \cite{DART_Engine}.

Command messages are published through the ROS2 API native to the PX4 flight stack. We use the latest (currently beta) version of the PX4 flight stack which utilizes the
$\mu\text{XRCE-DDS}$ middleware as the bridge between uORB and ROS2 topics. The Raspberry Pi receives position messages from an OptiTrack motion-capture system which are relayed as \textit{/fmu/out/vehicle\_odometry} messages to the flight controller.

In addition, we use an FrSky R9MX radio receiver connected to the Pixhawk board through SBus alongside an FrSky Taranis radio control transmitter to switch the quadrotor in and out of offboard control mode and for flight safety switches.

\begin{figure}[h!]
    \centering
    \includegraphics[width=0.2\textwidth]{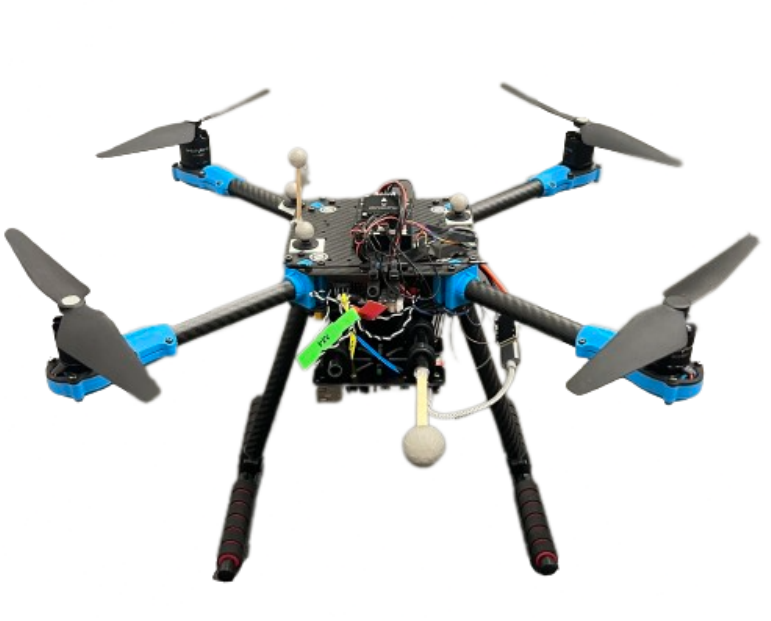}
    \caption{The Holybro x500v2 quadrotor used for hardware experiments. It is fitted with tracking markers, a radio receiver, and an on-board Raspberry Pi with ROS2 for control computations. The quadotor weighs \qty{1.69}{kg} and is \qty{0.5}{m} diagonally. }
    \label{fig:my_quadrotor_image}

\end{figure}\vspace{-1.5mm}
The ROS2 control algorithm node publishes the computed input, \emph{i.e.}, rate and thrust commands, at 100Hz. On average, over all experiments, the average computation time for the Newton-Raphson flow controller with I-CBF contraints is $7.138\times 10^{-5}\pm 2.971\times 10^{-5}$ seconds, well within the control update period. 

\subsection{Newton-Raphson Flow Controller Performance and Comparison}\label{subsec:experiments/flightdata}
In this section we compare the performance of our Newton-Raphson flow controller with a well-tuned PID-based baseline controller native to the PX4 flight stack \cite{PX4_Controller_Diagrams}.

\begin{figure}[h!]
    \centering
    \includegraphics[width=0.50\textwidth]{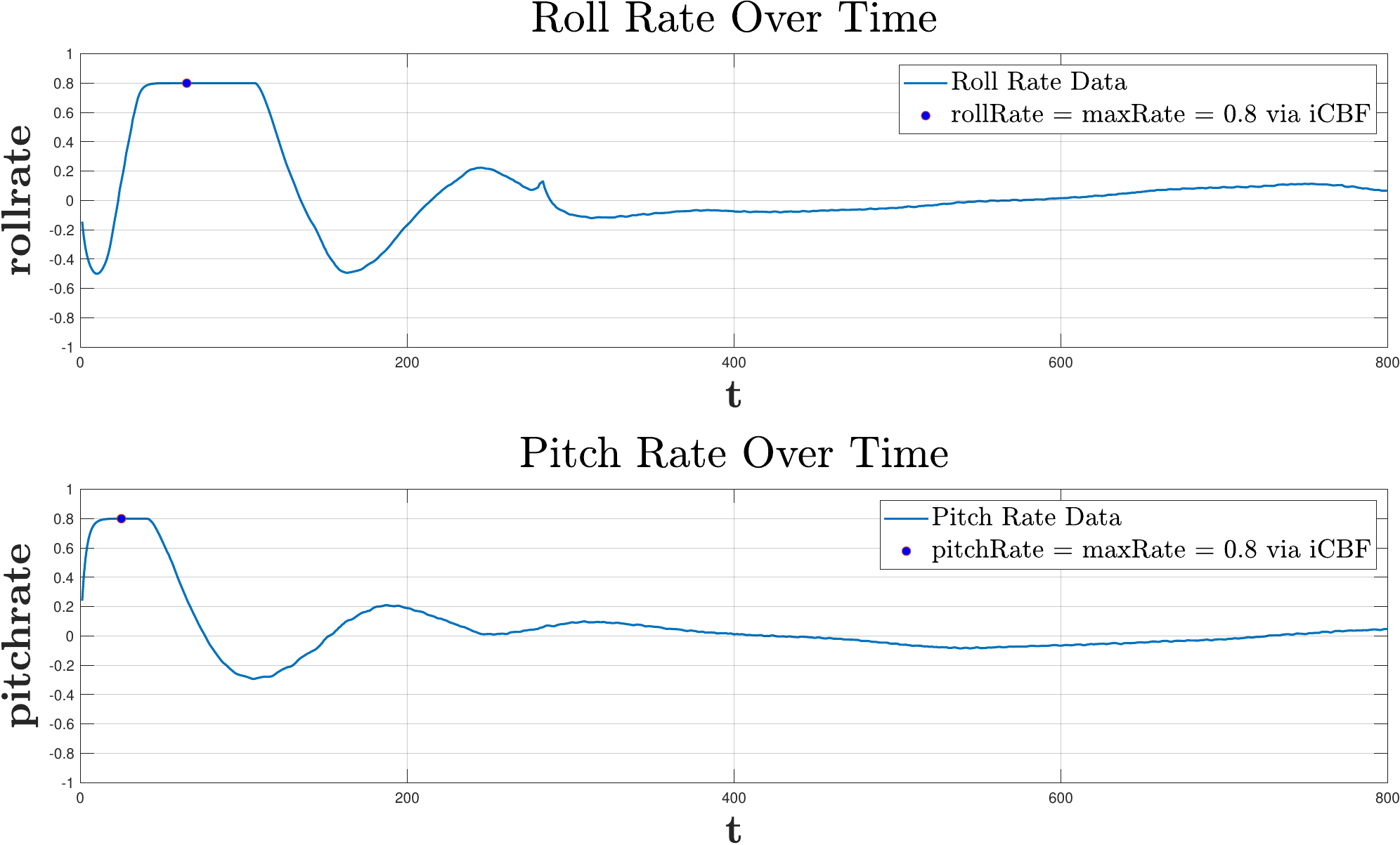}
    \caption{Roll and Pitch Rates Subject to Safety Constraints via Integral CBFs. The rates can be seen to be smoothly limited to a maximum absolute value of \qty{0.8}{rad/s}.}
    \label{fig:iCBFs}

\end{figure}

We test both controllers with five reference trajectories: vertical circle, horizontal circle, horizontal lemniscate, vertical tall lemniscate, and vertical short lemniscate.
The circular trajectories have periods of $3.14$ seconds while the lemniscates have periods of $6.28$ seconds. The results are shown in Figure \ref{fig:flight_trajecories_together}. The root-mean-squared error (RMSE) of the tracking trajectory is reported in Table \ref{tab:RMS_Comp}. We neglect the initial transient portion of flight for Figure \ref{fig:flight_trajecories_together} and Table \ref{tab:RMS_Comp} to focus our attention solely on the trajectory tracking portion of flight. In all cases, the Newton-Raphson flow controller outperforms the baseline controller. 
Some of the high-error trajectories for the baseline are due to the position of the quadrotor lagging behind the reference, which causes high error not necessarily evident in the plots of Figure \ref{fig:flight_trajecories_together}. Finally, Figure \ref{fig:iCBFs} shows an example of the effects of the I-CBF constraints. In particular, it is seen that the roll rate and pitch rates for the tracking of the horizontal circle are maintained within the $\pm 0.8 \ \text{rad}/\text{s}$ range of allowable values. 


\begin{table}[H] 
  \centering 
  \caption{Newton-Raphson vs Baseline RMSE}
  \label{tab:RMS_Comp}
  \begin{tabular}{|c|c|c|}
    \hline
    \textbf{Trajectory} & \textbf{Newton-Raphson} & \textbf{Baseline}\\
    \hline
    Vertical Circle & 0.051 & 0.26644\\
    \hline
    Horizontal Circle & 0.16834 & 0.61125\\
    \hline

    Horizontal Lemniscate & 0.15453 & 0.18775 \\
    \hline

    Vertical Short Lemniscate & 0.045291 & 0.14833 \\
    \hline

    Vertical Tall Lemniscate & 0.096611 & 0.22519 \\
    \hline

  \end{tabular}
\end{table}


\begin{figure}[h!]
    \centering
    \includegraphics[width=0.40\textwidth]{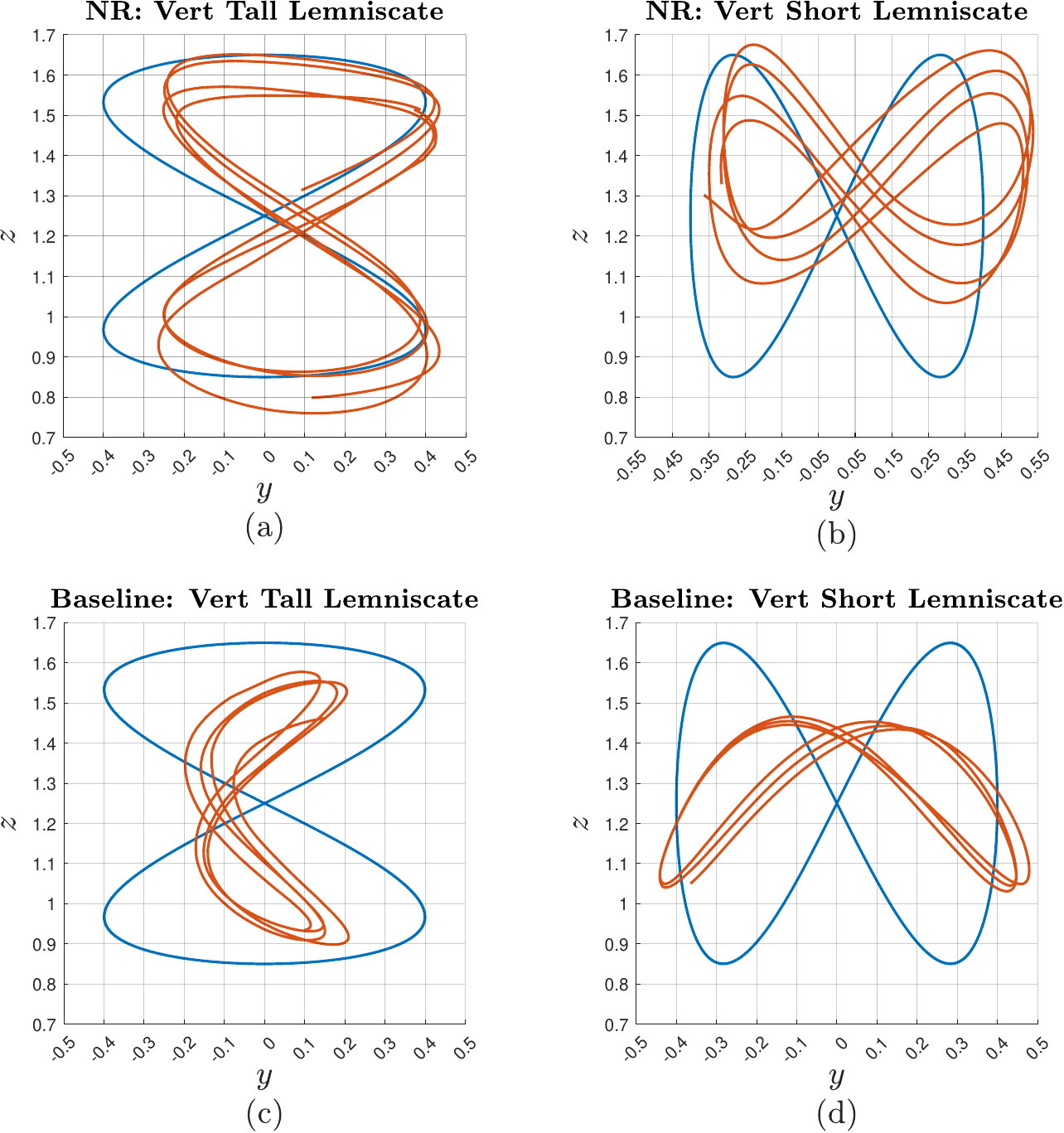}
    \caption{Comparison of Newton-Raphson controlller (top row) and Baseline controller (bottom row)  on fast vertical lemniscate trajectories with period $3.14$ seconds. The Newton-Raphson controller approximately maintains the lemniscate shape, while the baseline controller tracks poorly.}
    \label{fig:fast_trajectories}

\end{figure}

The efficacy  of the Newton-Raphson  controller is further exemplified when we speed up the lemniscate trajectories to be twice as fast with a period of $3.14$ seconds, as shown in Figure \ref{fig:fast_trajectories}. At this period, the tracking performance degrades for both the Newton-Raphson controller and the baseline controller, but the degradation of the baseline controller is severe while the Newton-Raphson controller is able to maintain the basic lemniscate shape. Table \ref{tab:RMS_Comp_fast}  displays the 
root-mean square errors for the applications of the two techniques to the two sped-up trajectories.


\begin{table}[H] 
  \centering 
  \caption{RMSE on Fast Trajectories}
  \label{tab:RMS_Comp_fast}
  \begin{tabular}{|c|c|c|}
    \hline
    \textbf{Trajectory} & \textbf{Newton-Raphson} & \textbf{Baseline}\\
    \hline

    Vertical Tall Lemniscate & 0.10508 & 0.27228 \\
    \hline

    Vertical Short Lemniscate & 0.12724 & 0.28747 \\
    \hline
    
  \end{tabular}
\end{table}

\section{Conclusion}\label{sec:conclusion}
This paper investigates the application of a tracking controller, recently defined and proposed by an author of this paper, to a quadrotor having to follow a prescribed flight trajectory. The tracking control in question is founded on the Newton-Raphson method for solving algebraic functions, output prediction, and a controller's speedup \cite{wardi_ijrnc_2019}. Thus far, the controller has been investigated by means of theory and simulation, and this paper describes its first application to a real system in a laboratory setting. We compare our proposed Newton-Raphson flow controller to the native, highly-tuned PID control of the PX4 autopilot stack.
 The Newton-Raphson controller outperforms this controller in terms of lower root-mean-squared tracking errors on all of the tested trajectories.

\appendix

\section{Full Matrices for Quadrotor Dynamics}\label{Appendix: Full Matrices}

The full matrices $A$ and $B$ used in the simplified predictor dynamics discussed in Section \ref{subsec:simplified predictor dynamics} are given by:

\[
A = \begin{pmatrix}
0 & 0 & 0 & 1 & 0 & 0 & 0 & 0 & 0 \\
0 & 0 & 0 & 0 & 1 & 0 & 0 & 0 & 0 \\
0 & 0 & 0 & 0 & 0 & 1 & 0 & 0 & 0 \\
0 & 0 & 0 & 0 & 0 & 0 & 0 & -g & 0 \\
0 & 0 & 0 & 0 & 0 & 0 & g & 0 & 0 \\
0 & 0 & 0 & 0 & 0 & 0 & 0 & 0 & 0 \\
0 & 0 & 0 & 0 & 0 & 0 & 0 & 0 & 0 \\
0 & 0 & 0 & 0 & 0 & 0 & 0 & 0 & 0 \\
0 & 0 & 0 & 0 & 0 & 0 & 0 & 0 & 0 \\
\end{pmatrix}
\]

\[
B = \begin{pmatrix}
0 & 0 & 0 & 0 \\
0 & 0 & 0 & 0 \\
0 & 0 & 0 & 0 \\
0 & 0 & 0 & 0 \\
0 & 0 & 0 & 0 \\
\frac{1}{m} & 0 & 0 & 0 \\
0 & 1 & 0 & 0 \\
0 & 0 & 1 & 0 \\
0 & 0 & 0 & 1 \\
\end{pmatrix}
\]

The full matrix $C$ used in the output calculation as discussed in \ref{sec:quadrotor model and predictor} is given by:

\[
C = \begin{pmatrix}
1 & 0 & 0 & 0 & 0 & 0 & 0 & 0 & 0 \\
0 & 1 & 0 & 0 & 0 & 0 & 0 & 0 & 0 \\
0 & 0 & 1 & 0 & 0 & 0 & 0 & 0 & 0 \\
0 & 0 & 0 & 0 & 0 & 0 & 0 & 0 & 1 \\
\end{pmatrix}
\]


\bibliography{99_refs} 
\bibliographystyle{ieeetr}

\end{document}